\documentclass[letterpaper, 10 pt, conference]{ieeeconf}  %

\IEEEoverridecommandlockouts                              %

\usepackage{times} %
\usepackage{amsmath} %
\usepackage{amssymb}  %
\usepackage{cite}
\usepackage{url}
\usepackage{graphicx}
\usepackage{float}
\usepackage{todonotes}
\usepackage{subcaption}
\usepackage{booktabs}
\usepackage{xcolor}
\usepackage{microtype}
\usepackage{hyperref}

\title{\LARGE \bf
Perspective-Shifted Neuro-Symbolic World Models:\protect\\A Framework for Socially-Aware Robot Navigation
}

\author{Kevin Alcedo$^{1*}$, Pedro U. Lima$^{1}$ and Rachid Alami$^{2}$%
\thanks{*corresponding author: 
{\tt\small kevin.alcedo@tecnico.ulisboa.pt}}%
\thanks{$^{1}$Institute for Systems and Robotics, Instituto Superior Técnico, Universidade de Lisboa, Lisbon, Portugal}%
\thanks{$^{2}$LAAS-CNRS, Artificial and Natural Intelligence Toulouse Institute (ANITI), Toulouse, France}%
}

\begin{document}

\maketitle
\thispagestyle{empty}
\pagestyle{empty}

\begin{abstract}
Navigating in environments alongside humans requires agents to reason under uncertainty and account for the beliefs and intentions of those around them. Under a sequential decision-making framework, egocentric navigation can naturally be represented as a Markov Decision Process (MDP). However, social navigation additionally requires reasoning about the hidden beliefs of others, inherently leading to a Partially Observable Markov Decision Process (POMDP), where agents lack direct access to others' mental states. Inspired by Theory of Mind and Epistemic Planning, we propose (1) a neuro-symbolic model-based reinforcement learning architecture for social navigation, addressing the challenge of belief tracking in partially observable environments; and (2) a perspective-shift operator for belief estimation, leveraging recent work on Influence-based Abstractions (IBA) in structured multi-agent settings. 
\end{abstract}

\section{INTRODUCTION \& BACKGROUND}
\label{section:intro}

A long-standing goal in robotics and artificial intelligence is the deployment of embodied intelligent agents capable of learning, reasoning, planning, and executing complex real-world tasks alongside humans. Effective deployment of mobile robots in human environments is contingent on their ability to navigate socially. Unlike in egocentric navigation, a socially navigating robot interacts with humans and other robots, aiming to achieve its navigation goals without degrading—and ideally improving—the experiences of other agents. Socially-aware robot navigation poses unique challenges as it merges the complexity of traditional robot navigation alongside moving humans, in addition to the behavioral and situational consequences of this participation. Beyond accomplishing their own goals, socially adept robots must also consider and accommodate the needs of others. Consequently, success in this problem requires endowing intelligent embodied agents with the ability to perform both socially-aware situation assessment and socially-aware behavior synthesis. Key research challenges include ensuring safety, social clarity, and adaptability to diverse interactions \cite{francis2023principlesguidelinesevaluatingsocial,Singamaneni2024}. Previous work addresses these through human preference alignment in reinforcement learning \cite{AligningHumanPreferences}, trajectory prediction \cite{liu2023intentionawarerobotcrowd}, formal verification methods \cite{FormalizingTrajectories}, symbolic human-aware representations \cite{shekhar2024humanaware}, and perspective-taking for joint human-robot planning and social signaling \cite{Khambhaita2017viewing,Singamaneni2021humanaware,Khambhaita2017headbody,Buisan2023ttc,Nikolaidis2016legibility}.

Social navigation requires agents to make real-time decisions while balancing safety, context, and social interactions, making reinforcement learning (RL) a compelling framework for training socially-aware policies that generalize across diverse interactions. Model-free RL methods rely on trial-and-error learning, often requiring millions of interactions to achieve high performance. In contrast, model-based reinforcement learning (MBRL) introduces an explicit model of the task, allowing agents to optimize policies with significantly fewer interactions; a crucial advantage for robotic applications where high-fidelity simulation may not be available, real-world data collection is costly, and/or exploration in the real platform can be unsafe or not possible.  However, the reliance on a model introduces challenges as models may be incomplete, imperfect, or computationally expensive to acquire and maintain.

To mitigate the limitations of hand-crafted or analytically derived models, data-driven statistical methods have emerged as a powerful alternative, offering surrogate models learned directly from data. For a comprehensive overview of relevant architecture designs to learn models from data, refer to  \cite{StateRepresentationOverview}. Recent advancements, often categorized under \textit{learning world models}\cite{WorldModels,RSSM,DreamerV1,DayDreamer}, have significantly improved the robustness of MBRL, making it a competitive alternative to model-free approaches with significant improvements in data-efficiency. Models in MBRL often assume fully observable environments or rely on approximate state representations. Mainly they are formulated within a Markov Decision Process (MDP) framework due to practical considerations such as computational complexity and scalability. Strategies such as frame-stacking, integrating recurrent networks or memory components, and learning latent state-spaces are commonly used to approximate Markovian state representations. While effective in many cases, they do not explicitly reason about uncertainty in state, which is crucial for social navigation tasks where an agent must predict and adapt to human behavior under incomplete information. Partially Observable Markov Decision Processes (POMDPs) provide a principled approach for modeling such uncertainty, enabling agents to maintain and update beliefs about unobservable aspects of the environment.

Given the complexity of social navigation, an ideal agent should not only learn from data but also encode structured reasoning about the task, social context, and other's behavior. Neuro-Symbolic architectures provide a promising approach by integrating learned representations with symbolic abstractions, enabling agents to infer and reason about latent states such as human beliefs and intent. Consequently, complex latent features learned through large-scale data can be distilled into more transparent symbolic forms, improving both reliability (through formal verification) and interpretability (through explicit formulas or constraints). By sharing a common symbolic language, human operators can confidently inject domain knowledge, thereby accelerating agent adaptation and ensuring alignment with task objectives. Furthermore, by modeling uncertainty explicitly in symbolic terms, these architectures can leverage existing symbolic solvers and planning algorithms used in safety/mission-critical scenarios. A recent systematic survey \cite{Acharya2023NeurosymbolicRL} presents a taxonomy of neuro-symbolic architectures applied to RL, highlighting diverse approaches to integrating learning and reasoning.

In particular, we seek to equip agents with the capability to reason, infer, and track others’ beliefs in social-navigation tasks. Under a sequential decision-making framework, egocentric navigation is naturally an MDP, but introducing hidden beliefs turns the problem into a POMDP where agents lack direct access to others' mental states. Drawing inspiration from cognitive science studies on Theory of Mind (ToM) and mental simulation \cite{Goldman2006}, we embed belief tracking in a neuro-symbolic MBRL architecture that adapts decisions to social context. ToM supplies the mental-state concept, and Epistemic Planning provides the logic for acting over beliefs \cite{Bolander2011EpistemicPF}. Guided by computational ToM models that cast belief inference as hidden variables in POMDPs \cite{BAYESIANINVPLAN_BAKER2011}, we restrict this study to first-order beliefs; while we do not yet exploit the full DEL guarantees of epistemic planners, our approach trades that formal rigor for data-driven scalability, and the symbolic latent factors could later interface with such formalisms. We therefore illustrate one way in which ToM and Epistemic Planning can co-manifest in a neuro-symbolic agent via latent-space mental simulation.

In the following sections, we formalize the problem of social navigation as a POMDP. We leverage recent work on the Influence-based Abstraction (IBA) framework and Influence-Augmented Local Models (IALMs) \cite{oliehoek2021}, enabling agents to integrate social reasoning into their decision process. We propose (1) a neuro-symbolic architecture that implements an IALM to learn belief-state representations from experiences; and (2) a social navigation mental simulation procedure, a perspective-shift operator, which functions as an algorithmic theory of mind in the context of social navigation. 

\begin{figure}
    \centering
    \includegraphics[width=\columnwidth]{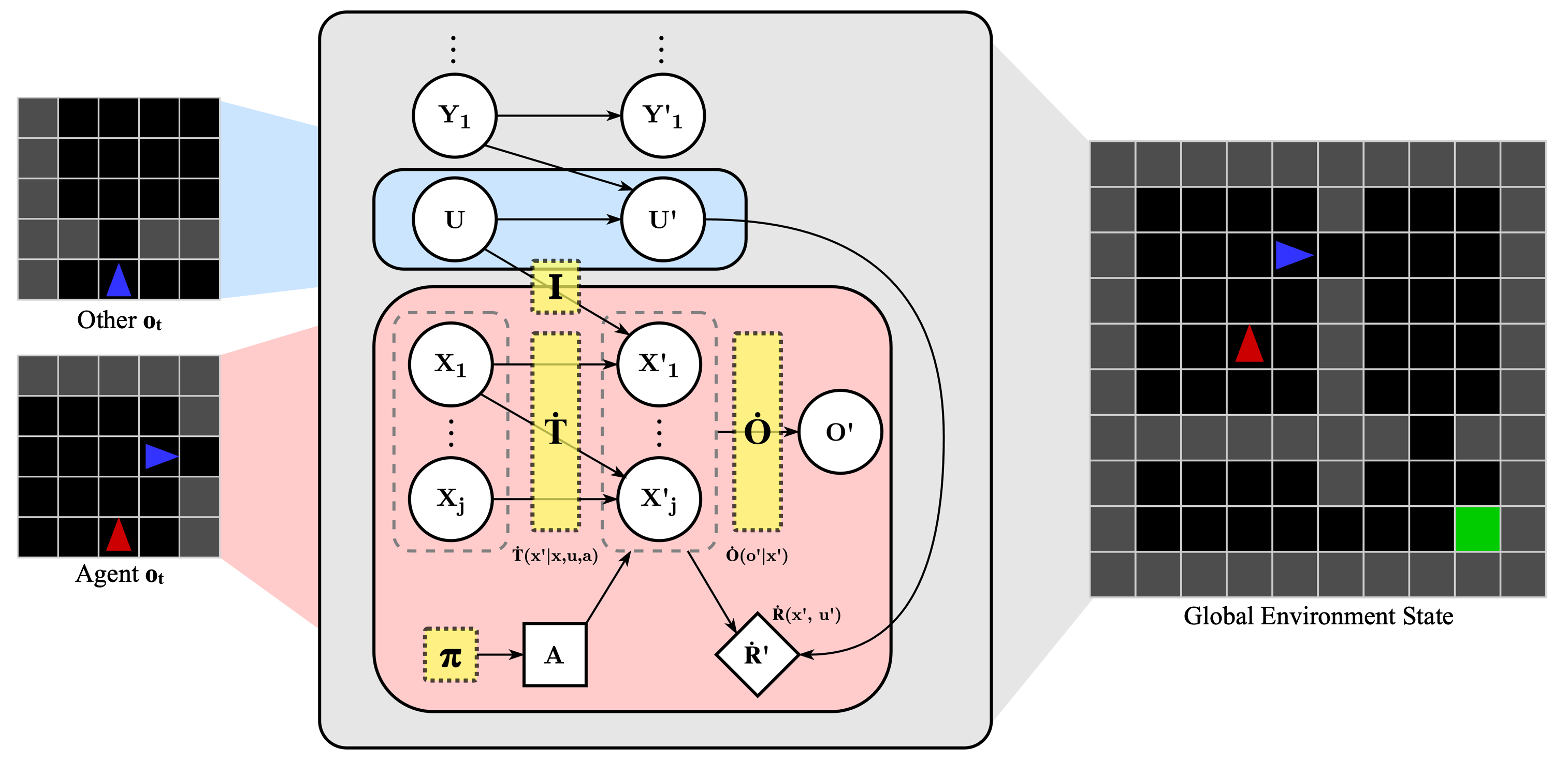}
    \caption{IALM for the person-following task.}
    \label{fig:environment}
\end{figure}

\begin{figure*}[t]
    \hfill
    \centering
    \subfloat[]{\includegraphics[width=0.8\columnwidth]{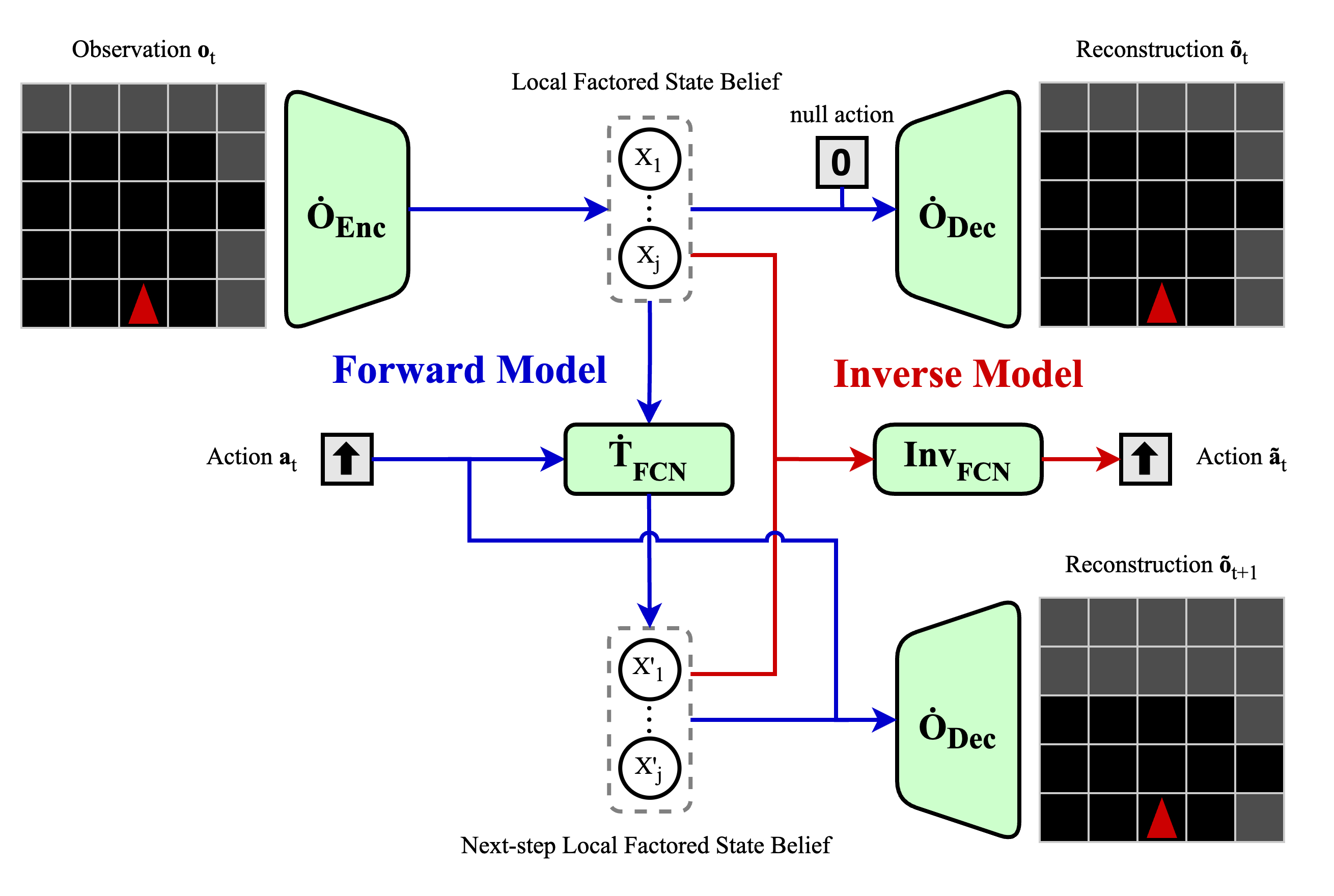}\label{fig:arch_a}}
    \hfill
    {\color{gray}\vrule width 0.5pt height 4cm}
    \hfill
    \subfloat[]{\includegraphics[width=0.6\columnwidth]{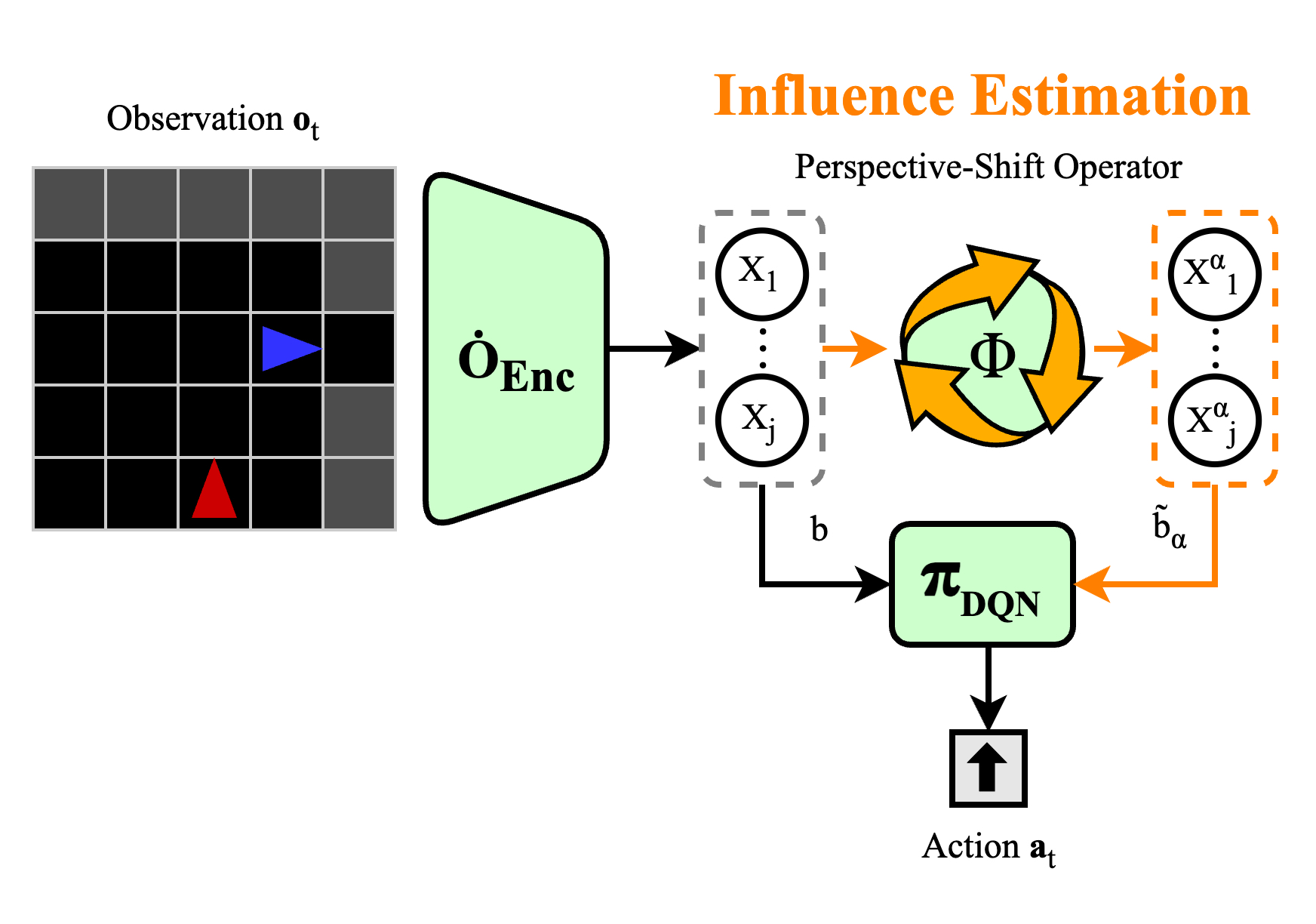}\label{fig:arch_b}}
    \hfill
    \label{fig:architecture}
     \caption{\textbf{(a) Neuro-Symbolic World Model:} A forward model (blue arrows) learns factored belief-state representations, regularized by action-relevant reconstructions from the observation decoder and an inverse model (red arrows). \textbf{(b) Influence-Augmented Policy:} A policy conditioned on the agent’s and perspective-shifted estimates of others’ beliefs (orange arrows).}
\end{figure*}

\section{PRELIMINARIES}

Next, we describe how the Influence-Based Abstraction (IBA) framework proposed by \cite{oliehoek2021} provides a principled methodology for formulating the socially-aware robot navigation problem. The main insight is that standard POMDP formulations require explicit modeling and belief tracking over the complete state-space, which becomes computationally unmanageable in large-scale multi-agent systems. While one could attempt to partition the state-space into independent localities for decision-making, complete partitioning is generally not possible due to conditional dependencies on other variables (e.g., other agents' states or environmental factors). IBA addresses this challenge by factorizing the global, partially observable environment into smaller, tractable local models, while identifying a set of influence points that compactly capture any remaining boundary dependencies. As shown in \cite{oliehoek2021}, these influence points enable construction of an Influence-Augmented Local Model (IALM), capturing global interactions through purely local computations—exactly so, if all relevant influences are accounted for. Although, in practice, identifying, modeling and/or tracking every influence point can be computationally prohibitive; requiring approximations such as those proposed by \cite{suau22a_ials}. The factorization done in IALM's  allows the problem to be modeled as a two-stage dynamic Bayesian network (2DBN) \cite{Poupart_FPOMDP,Boutilier1999_2TBN}, exploiting the conditional independence among non-influence variables. When influence points are modeled exactly they serve as a sufficient statistic (i.e., they capture all necessary information from non-local variables so that the local transitions no longer depend on the rest of the state), thus reducing computational complexity without sacrificing task performance. In socially-aware robot navigation tasks, one class of influence points naturally correspond to estimates of other agents' mental states, variables that are inherently unobservable yet crucial for effective decision-making.

Local Factorized POMDPs (Local-FPOMDPs) extends standard POMDPs by explicitly exploiting the structure of multi-agent environments through factorization. In a Local-FPOMDP, the global state space \(\mathcal{S}\) is decomposed into subsets of state variables, each subset corresponding to a local region that primarily interacts with the rest of the system through limited influence points. Because the environment is partially observable, the agent must maintain a belief over unobserved state variables. Rather than tracking a single monolithic belief distribution \(b(s)\) over the entire state \(s \in \mathcal{S}\), we factorize the belief according to the structure used for state-space decomposition. Specifically, we maintain a set of sub-distributions \(b(s_1), \dots, b(s_k)\), one for each factor. By factoring the belief, we track only the necessary interdependencies, which significantly reduces computational overhead and preserves essential factor couplings.

Formally, a Local-FPOMDP is a tuple  \mbox{\(\langle \mathcal{S}, S, A, T, R, \Omega, O, \gamma \rangle\)},  where \(\mathcal{S}\) is the global set of states, and  \mbox{\(S = \{S_1, \dots, S_k\}\)} the set of state factors, each \(S_i\) having domain \(\mathrm{dom}(S_i)\). Concretely, \mbox{\(\mathcal{S} \;=\; \times_{i=1}^k \mathrm{dom}(S_i) \;=\;\{\,\langle s_1,\dots,s_k\rangle \mid s_i \in \mathrm{dom}(S_i)\}\)}. Thus, each state \(s \in \mathcal{S}\) is given by a tuple  \(\langle s_1, \dots, s_k\rangle\),  where \(s_i \in \mathrm{dom}(S_i)\). Transitions follow \mbox{\(T(s, a, s') = P(s' \mid s, a)\)},  and the reward function $R(s,a)$ assigns a real-valued reward to each state--action pair. Because the environment is partially observable, observations come from  space \(\Omega\)  according to \(O(s', a, o) = P(o \mid s', a)\). Finally,  \(\gamma \in [0,1)\)  is the discount factor weighting future rewards.

\textit{Locality} arises by focusing on a subset of factors \mbox{$X = \{X_{1},\dots,X_{j}\} \subseteq S$} with  \mbox{$j \le k$} that is relevant to a particular agent or sub-problem, with local state space $\mathcal{X}$ defined by \mbox{$\mathcal{X} \;=\; \times_{i=1}^j \mathrm{dom}(X_i) \;=\;\{\,\langle x_1,\dots,x_j\rangle \mid x_i \in \mathrm{dom}(X_i)\}$}. Naively, one might attempt to track \mbox{$x = \langle x_1,\dots, x_j\rangle \in \mathcal{X}$} at each timestep by simply marginalizing out all other  variables in \(S\), but that is generally infeasible in large domains. Instead, IBA identifies a smaller set of non-local variables (influence sources) \(u \in U \subseteq S \setminus X\). This yields a local transition function  \mbox{\(\dot{T}(x' \mid x,u,a)\)}, a local observation function \mbox{\(\dot{O}(o \mid x, a)\)}, a local reward function \(\dot{R}(x,a,u)\) and an influence distribution \mbox{\(I(u \mid l)\)},  where \(l\) denotes the action-local-state history needed to infer the current value of \(u\). The  action-local-state history up to time \(t\) is \mbox{$l_t = \langle x_1, a_1, \dots, a_{t-1}, x_t\rangle$}, where \(x_t\) and \(a_t\) are the local state and action at time \(t\), respectively. Thus, \mbox{\(I(u_t \mid l_t)\)} captures the distribution over possible influences \(u\) at time \(t\), given the relevant local states and actions so far.

Consequently, the next local state is obtained by summing over those influence variables:
\[ P(x' \mid l, a) = \sum_{u} \dot{T}(x' \mid x, u, a)\,I(u \mid l) \] Here, the influence distribution \(I\) removes the requirement to track the entire global state; it compactly represents how non-local regions impact local transitions. In the context of a socially-aware robot, these influences may include unobserved beliefs or intentions of other agents.

By factorizing social tasks into local models with well-defined influence points, one can capture essential cross-agent interactions without exhaustive belief tracking over all global states (See Fig~\ref{fig:environment}). The resulting IALM provides a computationally tractable framework, as demonstrated in \cite{suau22a_ials}, while preserving accuracy in practice. 

\section{APPROACH}

Next, we describe our approach to learn factored belief-state representations and estimate the influence in order to construct an IALM applied to policy learning in socially-aware robot navigation tasks. 

\begin{figure*}[!ht]
    \centering
    \hfill
    \subfloat[]{\includegraphics[width=0.9\columnwidth]{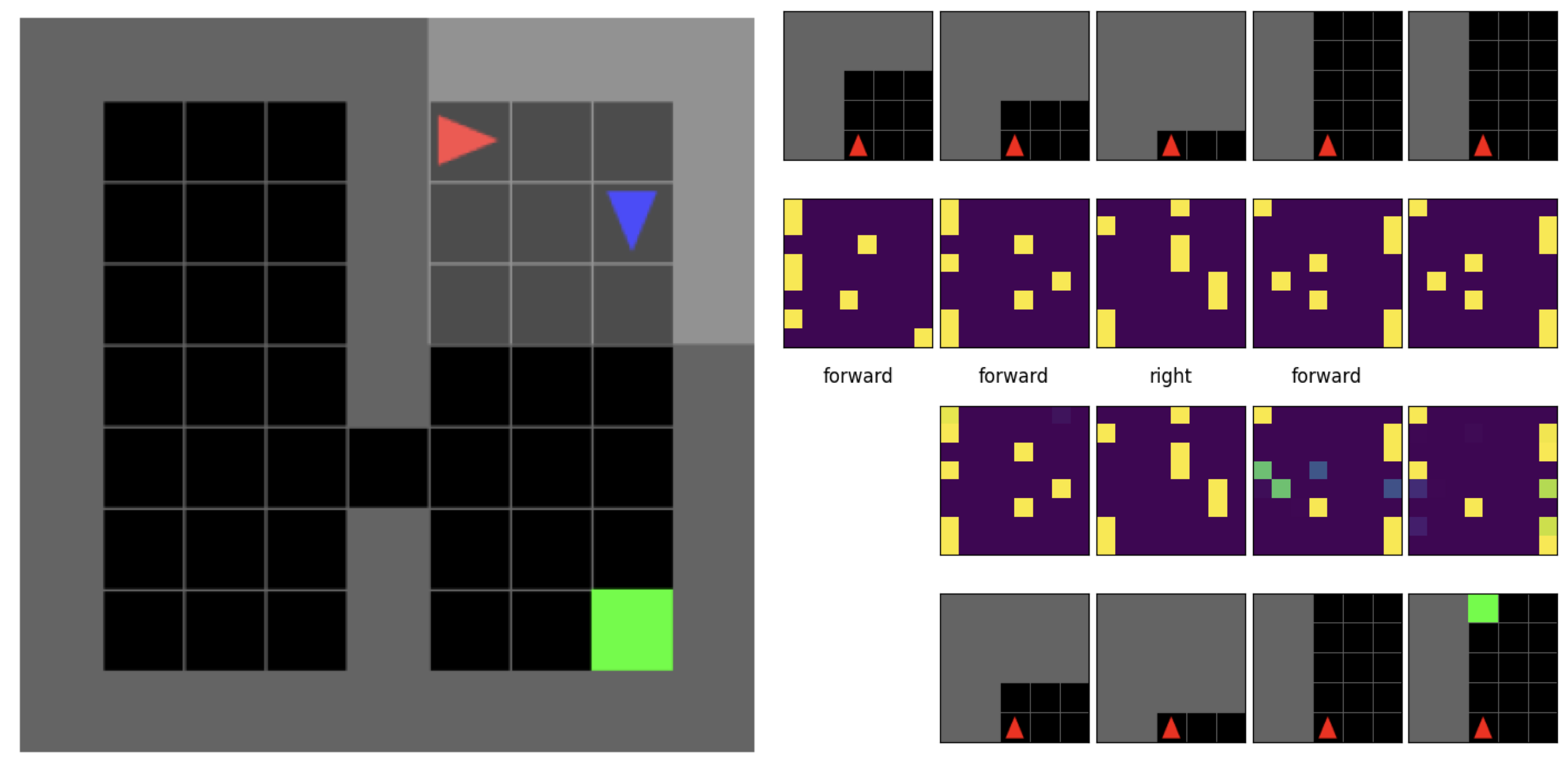}\label{fig:rep_00}}
    \hfill
    \subfloat[]{\includegraphics[width=0.9\columnwidth]{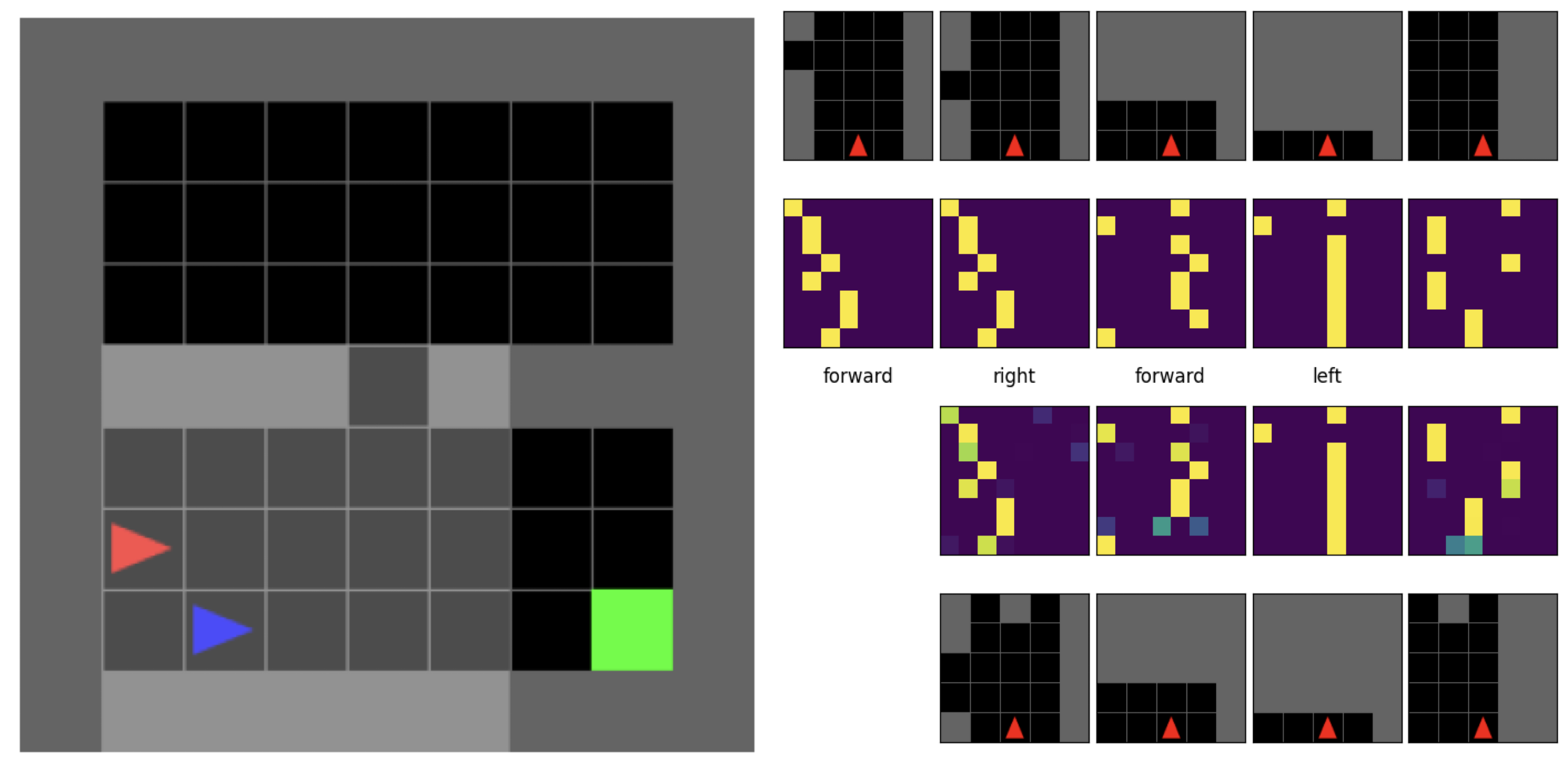}\label{fig:rep_01}}
     \caption{\textbf{World Model and Perspective-Shift:} Two illustrative examples demonstrating the agent's internal representation and belief updates as it performs a perspective-shift to estimate POI's beliefs. For each (a) and (b): \textbf{Left panel:} Global environment state ($9\times9$ map), indicating the agent (red), POI (blue) and goal (green) locations, with the agent's observable area ($5\times5$ view) in light-gray. \textbf{Right panel, top row:} Ground-truth observations from the agent's perspective. \textbf{Second row:} Factored belief-states, with rows representing distinct latent factors ($N=8$) and columns corresponding to discrete factor values ($K=8$); since each row corresponds to a belief it sums to one, where high values are bright yellow and low values purple. \textbf{Third row (text):} Imagined actions taken by the agent. \textbf{Fourth row:} Predicted perspective-shifted beliefs for each imagined action, inferred solely from the initial observation. \textbf{Bottom row:} Reconstructions of perspective-shifted beliefs.}
     \hfill
     \label{fig:representation_learning}
\end{figure*}

\subsection{Neuro-Symbolic World Model} 
\label{sec:neuro_symbolic_world_model}

As mentioned in Section~\ref{section:intro}, our interest is to explicitly model beliefs under the lens of MBRL in scenarios where we do not have access to $\dot{T}$ and $\dot{O}$, requiring us to learn them directly from experience in partially observable environments. To this end, we adopt a self-supervised Variational Autoencoder (VAE)\cite{Kingma2014} with a symbolic latent space implemented via the Gumbel-Softmax method \cite{GumbelSoftmax_01,GumbelSoftmax_02}. This architecture learns $N$ belief-state factors, each with $K$ discrete values, directly from raw observations. The Gumbel-Softmax reparameterization transforms these categorical variables into continuous relaxations, yielding a differentiable factored belief-state representation amenable to symbolic learning and reasoning. A fully-connected neural network (FCN) implements the local transition model, predicting next-step factors given a learned action embedding (see Figure~\ref{fig:arch_a}). To ensure these latent factors capture action-relevant information, we incorporate an inverse model \cite{StateRepresentationOverview}\cite{haimerl2025world} that infers the action from consecutive latent states. This additional signal regularizes the latent states to encode precisely those features crucial for decision-making and reduces ambiguity in partially observable settings. The loss function for our world model becomes: \[  \mathcal{L} = \mathcal{L}_{\mathrm{VAE}} + \mathcal{L}_{\mathrm{forward}} + \mathcal{L}_{\mathrm{inverse}} \] Where $\mathcal{L}_{\mathrm{VAE}}$ is the conventional loss function for the VAE, composed of the observation reconstruction error and Kullback-Leibler (KL) divergence between the learned factored belief posterior and uniform categorical prior; $\mathcal{L}_{\mathrm{forward}}$ is the KL divergence between predicted and observed next-step belief factors; and $\mathcal{L}_{\mathrm{inverse}}$ is the mean square-error (MSE) between the inferred and true action embedding\footnote{Because the model learns multiple representations simultaneously, we selectively stop gradient propagation to avoid interference in learning signals.}. Our proposed neuro-symbolic world model effectively learns a 2DBN where neural networks implement both the local transition and observation functions. Although extending a detailed symbolic integration is beyond the scope of this work, we envision these discrete latent factors facilitating future research with existing symbolic planners and interpretable decision-making methods in complex environments.

\subsection{Perspective-Shift Operator}
\label{sec:perspective_shift_operator}

In social navigation, estimating the beliefs of other agents often requires reasoning about their perspective. Equipped with a world model, an agent can simulate hypothetical trajectories and their associated belief updates without real interactions with the environment. Assuming other agents share similar sensing and motion capabilities, we posit that estimating others' beliefs in social navigation entails perspective-taking \cite{Flavell1992PerspectivesOP, PerspectiveTakingAlami}. Formally, we introduce the perspective-shift operator \(\Phi\):\[
\tilde{b}_\alpha = \Phi(b, \hat{x}_\alpha, \mathcal{M})
\]where \(b\) is the agent’s current (factored) belief, \(\hat{x}_\alpha\) is the locally observed state configuration of another agent \(\alpha\), and \(\mathcal{M} = (\dot{T}, \dot{O}, \dot{R})\) is the learned neuro-symbolic world model.

Intuitively, the operator \(\Phi\) leverages the local transition and observation models from \(\mathcal{M}\) to simulate how the world would appear if the agent occupied the state configuration represented by \(\hat{x}_\alpha\). Since we lack a global coordinate frame, the simulation treats the other agent’s configuration \(\hat{x}_\alpha\) as a local target state and updates beliefs through imagined navigation. The resulting belief \(\tilde{b}_\alpha\) encodes the observable environment from \(\alpha\)'s viewpoint, accounting for occlusions and partial observations that may differ significantly from the agent’s original perspective. The perspective-shift operator thus serves as a computational instantiation of Theory of Mind, providing a mechanism to estimate the influence distribution \mbox{\(I(u \mid l)\)} in the IALM formulation. Although this work targets social navigation, we envision applying \(\Phi\) to broader social‐reasoning tasks, e.g., collaborative assembly, where a perspective-shift would mentally simulate others' limb poses and the assembly step in progress.

\subsection{Influence-Augmented Policy Learning}

With a complete IALM representation---consisting of transition, observation, and influence models---we now optimize socially-aware navigation policies using reinforcement learning. We employ a Deep Q-Network (DQN) \cite{mnih2015humanlevel} to approximate the value function over the structured belief states produced by our neuro-symbolic world model. The inputs to our DQN are the current factored beliefs: both the agent’s own belief $b$ and the estimated belief of others $\tilde{b}_\alpha$. The network outputs Q-values corresponding to a discrete set of motion primitives. The  architecture is depicted in Figure~\ref{fig:arch_b}.

By directly providing structured belief states as inputs, our approach effectively approximates a value function defined over a belief-MDP \cite{KaelblingCassandra1998}, aligning with the POMDP formulation. This differs fundamentally from traditional uses of DQN in model-free settings, where values are learned from raw observations. Combining symbolic belief-state representations produced by the neuro-symbolic world model with value-based policy learning bridges classical belief-MDP ideas and scalable deep RL, while reducing the computational complexity associated with belief updates. 
We adopt DQN because its value-based nature matches our discrete action space and aligns with classical value-based POMDP solvers, enabling direct future comparisons. Finally, explicit belief modeling opens avenues for integrating our neuro-symbolic world model with traditional belief-space planners and symbolic reasoning methods.

\section{EXPERIMENTS \& RESULTS}

We selected person-following as a representative social navigation task that arises in real-world settings, such as service mobile robots operating in home environments, hospitals, or retail spaces. In such scenarios, robots are frequently expected to navigate socially; following and assisting humans to destinations or providing supportive behaviors. Person-following encompass key challenges in social navigation: reasoning about human intent, predicting future trajectories, handling partial observability, and maintaining socially acceptable distances (proxemics) and behaviors.

\subsection{Environment \& Task}

\paragraph{Environment} We use Minigrid~\cite{MinigridMiniworld23}, a partially observable gridworld with discrete observations, states and actions. The environment state encodes objects and their attributes in a 2D grid, while the agent observes a local region directly ahead. The agent can move, rotate, and interact with objects, receiving rewards based on its actions. Minigrid supports predefined scenarios but lacks multi-agent or social navigation tasks; thus, we introduce a custom scenario described next.

\paragraph{Person-Following Task} In our task (see Fig.~\ref{fig:environment}), the agent’s objective is to efficiently follow a person-of-interest (POI) to an unknown goal without a predefined map, while adhering to proxemics and avoiding collisions. At each timestep, the agent receives positive reward for perceiving and staying near the POI, and negative reward for collisions. An additional large positive reward is given when the POI reaches the goal, terminating the episode\footnote{Episodes are truncated if the agent strays too far from POI or exceeds a maximum number of steps. Full environment, task, and reward specifications are available at \url{github.com/alcedok/SocialNav_ROMAN25}.}. To better align with realistic social navigation constraints, future implementations could explicitly encourage proxemics aligned with human preferences or promote clear visibility to the human.

\paragraph{Experience Collection} The world model is trained on experiences collected using a mixture of random and expert policies $\{\pi_{\mathrm{random}}, \pi_{\mathrm{expert}}\}$; where $\pi_{\mathrm{expert}}$ implements $A^{*}$ planning using the global state. Episodes generated by the random policy ($50\%$) ensure diversity and exploration, while the expert episodes ($50\%$) illustrate optimal trajectories. In real-world scenarios, such expert policies could be gathered from human demonstrations or teleoperation data. Our experiments use a total of $3,000$ episodes of randomly generated maps (${\sim}300,000$ steps total).

\subsection{Model Training and Experiment Setup}

\paragraph{Model Training} Training proceeds in two stages: (1)~representation learning, where the world model (implemented as a VAE with convolutional encoder and transposed convolutional decoder, with self-attention at the middle layers) is trained from previously collected experiences (Figure~\ref{fig:arch_a}); and (2)~policy learning, where the DQN is trained on the person-following task. In stage (2), inputs to the DQN include the agent's own factored beliefs and the POI's estimated beliefs computed via the perspective-shift operator when the POI is visible, otherwise defaulting to a uniform random distribution (Figure~\ref{fig:arch_b}).

\paragraph{Setup} To evaluate our proposed perspective-shifted world model approach, we compare it against two reference conditions: (1)~a baseline model where the influences are sampled from a uniform random distribution (lower bound), and (2)~a perfect-information scenario (upper bound). Intuitively, a noisy random estimate of POI’s internal state should yield poorer performance, while having direct access to it should achieve near-optimal results. We train a single world model shared across all experiments.

\begin{figure}[h]
    \centering
    \includegraphics[width=0.7\columnwidth]{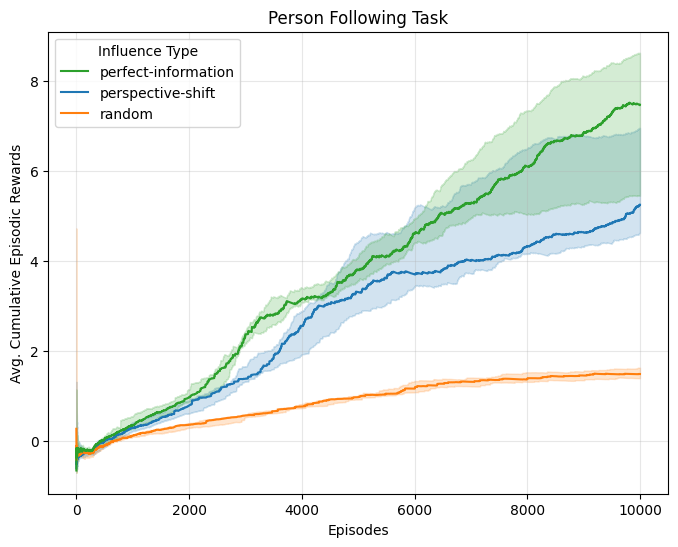}
    \caption{\textbf{Reinforcement Learning:} Influence-augmented policy learning curves for three influence estimation methods in the person-following task: perfect-information (upper-bound baseline), random noise (lower-bound baseline), and our proposed perspective-shift operator.}
    \label{fig:reinforcement-results-0}
\end{figure}

\subsection{Results}

\paragraph{Representation Learning} Figure~\ref{fig:representation_learning} illustrates the representations learned by the world model and predictions by the perspective-shift operator.  Notably, increased uncertainty after certain actions is reflected by noisier or ambiguous predictions (fourth row), and the agent occasionally hallucinates goals or produces innacurate reconstructions, i.e. misplaced walls (bottom row). Nevertheless, the agent's capacity to predict perspective-shifted beliefs highlights promising reasoning capabilities, warranting further analysis in future work.

\paragraph{Reinforcement Learning} Figure~\ref{fig:reinforcement-results-0} summarizes policy performance across influence types. Curves show average cumulative episodic rewards across $5$ independently trained policies, with shaded regions indicating bootstrapped $95\%$ confidence intervals~\cite{rliable_agarwal2021deep}. As hypothesized, policies using our perspective-shift operator substantially outperform the random baseline, demonstrating the value of explicitly modeling others' beliefs under partial observability. Our approach thus effectively bridges the performance gap toward the perfect-information upper bound.

\section{CONCLUSION \& FUTURE WORK}
Our neuro-symbolic MBRL approach formulates social navigation as a POMDP by leveraging IALMs to factorize the environment, significantly reducing complexity in belief modeling and tracking in multi-agent settings. By incorporating a perspective-shift operator inspired by Theory of Mind and Epistemic Planning, agents explicitly estimate others' beliefs through mental simulation. Experimental results in a person-following task demonstrate our method's potential for effective social reasoning, highlighting promising directions for extending the approach to richer environments with enhanced semantics, nuanced social dynamics, fully symbolic reasoning, and traditional belief-space planning methods.

\addtolength{\textheight}{0cm}   %

\section*{ACKNOWLEDGMENT}
\vspace{-0.15cm}
We thank Caroline Haimerl for insightful discussions on inverse models in neuroscience, and \mbox{David Rosenbluth} for discussions on neuro-symbolic architectures. This work was supported by LARSyS FCT funding (DOI: 10.54499/LA/P/0083/2020, 10.54499/UIDP/50009/2020, and 10.54499/UIDB/50009/2020) and the Portuguese Recovery and Resilience Plan (PRR) through project C645008882-00000055, Center for Responsible AI; and partially supported  by ANR (Agence Nationale de la Recherche) AAPG-2024/Ostensive project.
\vspace{-0.3cm}

\bibliography{IEEEabrv, bibtext}

\end{document}